\documentclass{article}

\usepackage[preprint,nonatbib]{neurips_2021}

\usepackage[numbers]{natbib}

\usepackage[utf8]{inputenc} % allow utf-8 input
\usepackage[T1]{fontenc}    % use 8-bit T1 fonts
\usepackage{url}            % simple URL typesetting
\usepackage{wrapfig,lipsum,booktabs}
\usepackage{amsfonts}       % blackboard math symbols
\usepackage{nicefrac}       % compact symbols for 1/2, etc.
\usepackage{microtype}      % microtypography
\usepackage{graphicx}
\usepackage{xcolor}         % colors
\usepackage{caption}
\usepackage{subcaption}
\usepackage{floatrow}
\usepackage[]{authblk}

\usepackage{amsmath}
\usepackage[ruled,vlined]{algorithm2e}
\floatname{algorithm}{Procedure}
\title{PatchNet: Unsupervised Object Discovery based on Patch Embedding}

\author[1]{Hankyu Moon}
\author[2]{Heng Hao}
\author[3]{Sima Didari}
\author[4]{Jae Oh Woo}
\author[5]{Patrick Bangert}
\affil[1,2,3,4,5]{Samsung SDS Research America}
\affil[ ]{\texttt{\{hankyu.m,h.heng,s.didari.jaeoh.w,p.bangert\}@samsung.com}}

\begin{document}

\maketitle

\begin{abstract}
We demonstrate that frequently appearing objects can be discovered by training randomly sampled patches from a small number of images (100 to 200) by self-supervision. Key to this approach is the {\it{pattern space}}, a latent space of patterns that represents all possible sub-images of the given image data. The distance structure in the pattern space captures the co-occurrence of patterns due to the frequent objects. The pattern space embedding is learned by minimizing the contrastive loss between randomly generated adjacent patches. To prevent the embedding from learning the background, we modulate the contrastive loss by color-based object saliency and background dissimilarity. The learned distance structure serves as {\it{object memory}}, and the frequent objects are simply discovered by clustering the pattern vectors from the random patches sampled for inference. Our image representation based on image patches naturally handles the position and scale invariance property that is crucial to multi-object discovery. The method has been proven surprisingly effective, and successfully applied to finding multiple human faces and bodies from natural images.
\end{abstract}

\section{Introduction}
The main challenge of object detection stems from (1) the variations of location, size, and pose of objects, (2) the occurrence of multiple objects in the same scene with possible occlusions, and (3) the infinite diversity of backgrounds. Modern CNN technology is capable of handling these challenges by its design of convolutional architecture that enables position and scale invariance and by fitting millions of neural network parameters. Both the network architectures and the sheer number of trainable parameters provide enough capacity to identify objects under diverse scene contexts and imaging conditions. The current state of the art object detection approaches employ either ‘proposal’ based \cite{wang2017fast} or ‘one shot’ based \cite{redmon2016you,liu2016ssd} approaches over deep features learned by CNNs. They are quite capable of finding objects and classifying them into hundreds or even thousands of categories.

Despite the progresses and their real-world utility, there is a major limitation with the current approaches: training a CNN for object detection requires a large number of images with object boundaries manually marked. This common limitation naturally motivates the needs for ‘unsupervised’ or ‘semi-supervised’ frameworks. There have been recent developments in unsupervised image representation learning and clustering that show promises toward unsupervised classification. Most notably, deep clustering based on pre-text training \cite{Ji_2019_ICCV,van2020scan} is approaching classification accuracy closer to the supervised state of art. This class of approaches is able to find clusters of images that can map to one of the specified object categories without supervision. This is enabled by the combination of clustering in neural embedded feature space, and self-supervised contrastive learning that constrains the embedding. Despite their surprising effectiveness toward classification tasks, the deep clustering approach in its current form is not applicable to detection problems due to the unconstrained range of possible object scales and positions.

For this type of unsupervised approach to be effective, each image should have a single object properly cropped so that it is close to the image center and with limited size variations. This conventional idea of  ‘an object as a vector’ inherently lacks the key ability to (1) represent multiple objects (while preserving their locations and sizes) in the image, and at the same time (2) extract out these objects into a common mathematical representation with which they can be compared and analyzed. The second condition clearly requires the objects to be aligned. This is especially critical because misalignment in both position and scale incurs highly nonlinear changes in pixel space. Obviously, availability of tight bounding boxes around each object will satisfy these two key requirements. Under the unsupervised scenario, however, this is essentially a chicken-and-egg problem.
\begin{figure}
     \centering
     \subfloat{\includegraphics[width=.99\linewidth]{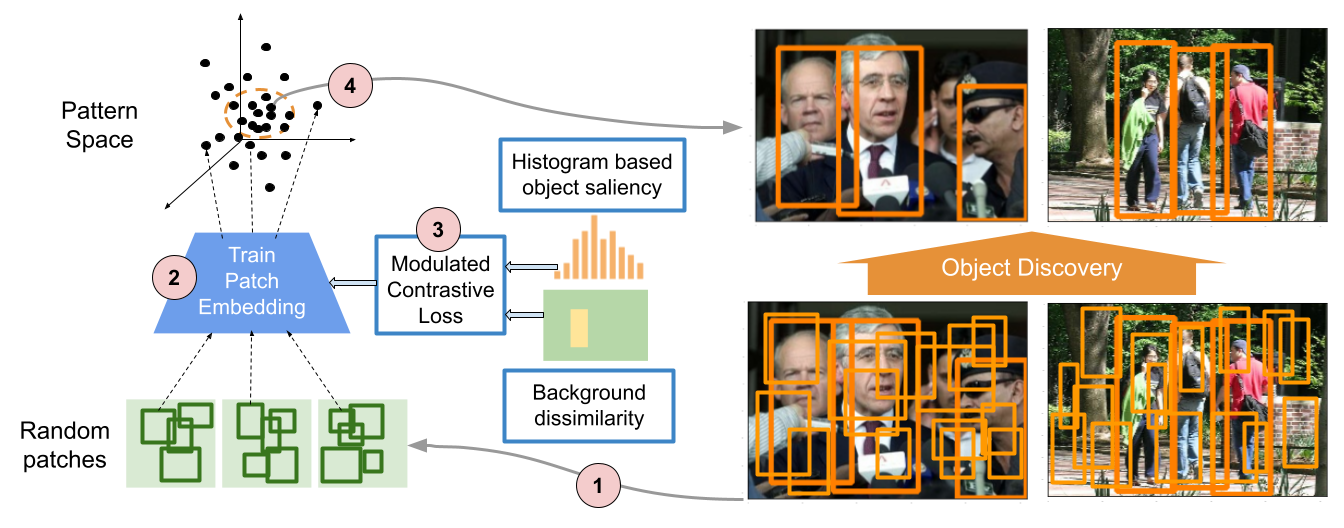}}
     \caption{Overall scheme of the approach }
     \label{fig:concept}
\end{figure}

In our proposed work, we utilize a patch-based image representation that fulfills the critical needs toward unsupervised object finding mentioned above. The patches sample all possible sub-images in a given image set, and are capable of capturing any number of objects. We combine self-supervised training with object saliency constraints that learns the mapping from the set of all sub-images to a common latent space. The learned mapping is tuned to patterns that frequently appear in the image set, and sends these patterns to a tight cluster. This self-emergent structure in the pattern space corresponds to the frequently appearing objects. Figure \ref{fig:concept} illustrates our overall scheme with numbered items below point to the circled numbers in the figure: 
\begin{enumerate}
\itemsep0em 
\item Image patches sampled at random locations and scales over an image set,
\item A neural embedding of these patches into a {\it{pattern space}} learned by contrastive training,
\item {\it{Objectness}} constraints by modulating the contrastive loss during training, so that the learned embedding is sensitive to object saliency while suppressing the background patterns commonly occurring in the image set, and
\item An unsupervised clustering step that identifies the frequently appearing objects in the images that leads to object discovery.
\end{enumerate}

Our approach is completely label-free because our embedding networkis  trained from random weights. We primarily show applications to the problem of person detection. This is in contrast with the existing Object Discovery work \cite{Cho2015,Vo2019,Vo2020} that shows success in discovering objects such as airplanes, cars, animals having typical color/texture and appearing against simple backgrounds (sky, pavement, grass, etc.). These proposal based approaches suffer from the {\it{part domination}} issue when applied to human detection due to diversity in clothing and background. We show that our approach is able to train a convolutional encoder network with a very small number of images (200 or less) and discover human bodies and faces at high success rates. We suspect this is due to the highly efficient patch based image representation that learns the {\it{building blocks}} of objects , in contrast to the whole-image based approaches that require a larger number of images to learn diverse object instances under differing imaging conditions.

\section{Related work}
\label{relatedwork}
Several approaches have emerged for object detection that require fewer ground truth bounding boxes or only image-level labels such as semi-supervised, weakly-supervised, and object colocalization models. Multiple Instance Learning (MIL) \cite{Bilen2016} and its modified variations are one of the main approaches developed for weakly supervised problems. MIL assumes that each image consists of multiple instances or bags of regions. A bag is called positive if it contains an object and MILs models try to find regions in the positive bags. Several studies were conducted to improve the performance of MIL approaches by enhancing the region proposal part, increasing the robustness of optimization, solving part domination issues, and memory efficiency \cite{Ren2020}. Attention and recurrent modules were developed and integrated with MILs to refine the region proposal selection tasks which were originally based on standard proposal generation methods such as selective search \cite{Tang2017,Tang2018,Li2019}. 
Weakly supervised or semi-supervised models require the training of a subset of training data with class level information. To alleviate the need for classification annotation of the training dataset, the colocalization methods bring us one step closer to the fully unsupervised learning. In these approaches either a pretrained CNN model (on classification tasks) or an off-the-shelf region proposal model is used to localize and detect the objects. Cho et al \cite{Cho2015} proposed a two-part colocalization model consisting of a region proposal matching and object localization steps. Randomized Prim’s algorithm \cite{Manen2013} generates proposal regions. Then region pairs are compared and ranked by Probabilistic Hough Matching (PHM) algorithm. To further improve the performance of the aforementioned approach \cite{Cho2015} Vo et al \cite{Vo2019} introduced a graph representation for capturing the similarity of the image pairs and formulated as a graph optimization problem. They also leveraged features learned from a pretrained CNN model \cite{Vo2020} for region proposal. These modifications lead to better performance for capturing multiple objects. 
\\ In a departure from two-part localization methods MONet\cite{monet2019}, IODINE \cite{IODINE2019} and GENESIS \cite{GENESIS2019} models were developed that decomposed scenes into objects using sets of latent variables for an object-based representations. To further improve these set-structured based object discovery models \cite{Locatello2020} proposed a Slot Attention module that learns the set-structured latent space. It was assumed that each set element corresponds to an object within the image. Hence, feeding learned features from a pretrained CNN model to the Slot Attention module coupled with a decoder, the objects can be localized and detected. Wei et al \cite{Wei2019} proposed Deep Descriptor Transformation (DDT) that used the correlation between the features learned by a pretrained CNN network for the images in dataset. Rambhatla et al \cite{rambhatla2021} proposed a dual memory model that uses the prior knowledge gained from detection of labeled categories to discover the unseen categories in a new unlabeled set. Zhang \cite{Zhang2020} proposed the Object Location Mining (OLM) model. OLM used feature representation of pretrained CNNs and salient pattern mining for object localization. The activation map generated by a pretrained CNN model (such as VGG-16) for each feature maps was used as an indicator to object existence. By extracting the activation maps and their connectivity and frequency, the salient objects were discovered.
Our patch based representation and training is unique in the sense that it is purely unsupervised and does not depend on any object proposal, either neural network based \cite{Manen2013} or deterministic feature based \cite{uijlings2013selective}. To the best of our knowledge, there have not been any other successful approaches that are able to discover multiple pedestrians from complex scenes.

\begin{figure}
     \centering
     \subfloat{\includegraphics[width=.95\linewidth]{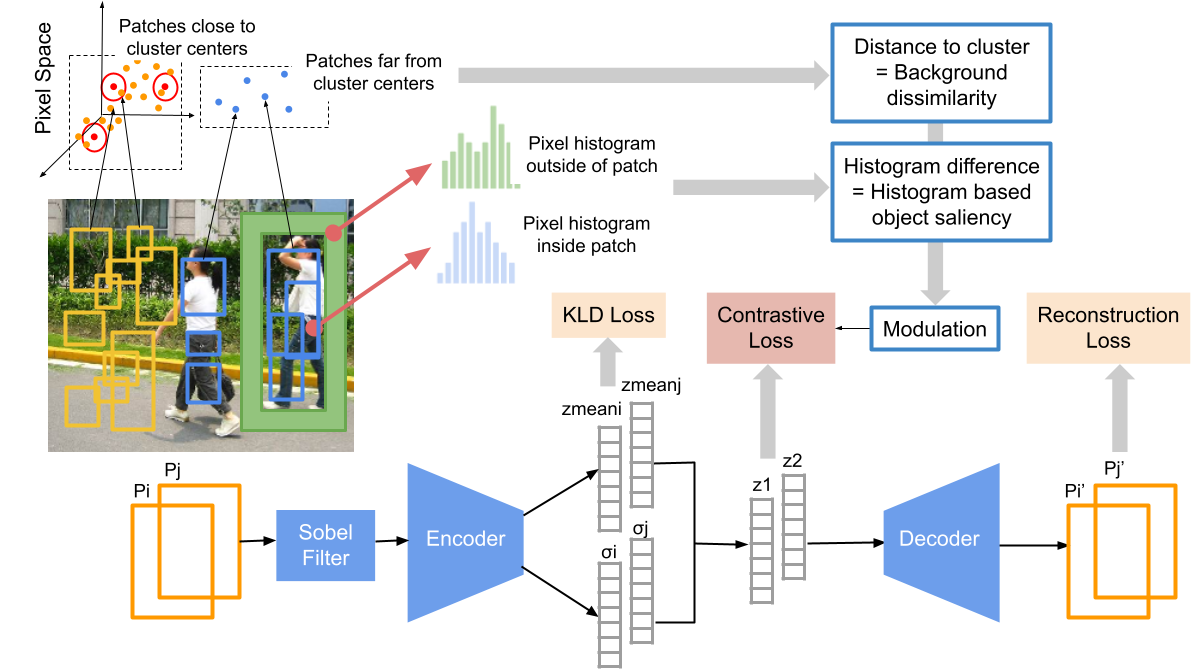}}
     \caption{Training architecture and contrastive loss modulation}
     \label{fig:architecture_modulation}
\end{figure}
\section{Patch space representation}
\label{patchgraph}

The patch based image representation satisfies the following properties, leading to unsupervised object discovery: (1) sampled image patches at multiple image positions and scales naturally satisfies position and scale invariance, (2) the learned association between the patterns from neighboring patches captures common object signatures, and (3) the patch based image representation is more efficient in the sense that it captures local features that are common to objects of the same or different classes. The existing approach of representing the whole object using a single bounding box and learning these instances in the whole image requires a very large number of samples.

Patches are sampled from an image frame at random positions and scales. The ranges of the size and aspect ratio (height divided by width) of the patches are determined such that the patches can capture objects having different scales and aspect ratios. The scale and aspect ratio of each patch are uniformly sampled from these ranges: $\left[ \text{scale}_{\text{min}}, \text{scale}_{\text{max}} \right]$ and $\left[ \text{ratio}_{\text{min}}, \text{ratio}_{\text{max}}\right]$. (Scale sampling scheme here and details in the Appendix) In our experiments, the scale range is fixed at $[20, 256]$ for images scaled to $\text{width}=256$. The aspect ratio range is a hyper-parameter and prior knowledge about object shape is used. We found that the aspect ratio of $\text{ratio}_{\text{min}} = \text{ratio}_{\text{max}} = 3.0$ for person discovery and $1.67-2.0$ for face + upper body discovery are found suitable. 
The patches are generated in pairs so that the local pattern association can be learned during training; each pair is generated so that the two patches overlap (high Intersection over Union). More specifically, pairs of patches $\left( p_1^{i}, p_2^{i} \right), i=1,\cdots, B$ are sampled as an input batch (of size $B$) such that $\text{IoU}\left( p_1^{i}, p_2^{i} \right) > 0.75$.

\section{Patch embedding}
The sampled patches are mapped to high-dimensional pattern vectors such that (1) each vector encodes the pattern information inside the patch and (2) the pairwise distance between the vectors reflects the co-occurrence of the corresponding patterns in natural images. The second property is enforced by the self-supervised training with a contrastive loss. The contrastive loss attracts adjacent pairs of patches together in the mapped latent space, while pushing unrelated pairs apart. At the end, the large number of embedded patches sampled from the image are endowed with a local distance structure. This is a 3D manifold of 2D position and scale in the high dimensional latent space where the {\it{bending and twisting}} of the 3D manifold attains the memory of the observed co-occurrences of patterns due to object presence. We add additional constraints to the contrastive training, so that the learned distance structure in the latent space captures the pattern correlation that originates from the objects to be discovered. The details will be provided in Section \ref{subsec:objectness}. We call this latent space a {\it{Pattern Space}} that represents all possible patterns in the given image set and their {\it{object-induced associations}}. 

Another feature of our approach is that diverse backgrounds facilitate estimating correct object boundaries. Our proposed training captures the consistent pattern co-occurrence inside frequent objects, while ignoring accidental co-occurrences due to backgrounds. In typical object recognition problems, the background diversity poses a challenge. In our case, the diversity actually helps by amplifying the correlation within objects and suppressing the accidental correlation between the objects and their background.

\subsection{Learning the patch embedding}\label{subsec:patch_embedding}
We train a Variational Autoencoder (VAE) \cite{kingma2013auto} with convolutional layers to learn the embedding, and its architecture is shown in Figure \ref{fig:architecture_modulation}. We employ a ResNet-18 backbone plus a global average pooling layer and a fully connected layer generating the latent pattern vector. It is trained from randomly initialized weights. The inputs are the sampled pairs of patches $\left( p_1^{i}, p_2^{i} \right)$ resized to a standard size of 32x32. We apply the Sobel filter to the 32x32 patches and feed the gradient images (2 channel $\left( dx, dy \right)$ images) to the training. This serves to reduce the variability of the image patterns, as found effective in \cite{caron2018deep,Ji_2019_ICCV}. The estimated mean and standard deviation $\left( z_{\text{mean}}^i, \sigma^i \right) =: \text{Encoder}(p^i)$ of the pattern vectors are 100-dimensional each. As in standard VAE training, latent samples are generated by $z^i \sim  N\left( z_{\text{mean}}^i, \sigma^i \right)$, and are fed to the decoder: $\text{Decoder}(z^i) := \left({p^i}\right)'$ that reconstructs the original patch. The decoder consists of two fully connected layers and four deconvolution layers.

The training loss breaks down to three different losses: (1) contrastive loss, (2) reconstruction loss, and (3) variational KLD loss as used in VAE \cite{kingma2013auto}.
As described in the previous subsection, the contrastive loss is the main driver of the self-supervised training. Let $[B]=\{1,\cdots,B\}$ where $B$ is the number of images in a batch. The highly overlapping pairs $\left( p_1^{i}, p_2^{i} \right), i\in [B]$ are the positive examples and the non-overlapping pairs $\left( p_1^{i}, p_2^{j} \right)$ for $i\neq j$ and $i,j\in [B]$ from different images are the negative examples. The standard Noise Contrastive Estimator (NCE) loss is used to pull matching $\left( \text{IoU} > 0.75 \right)$ pairs close together while pushing non-patching pairs apart.
 
 The NCE loss is designed to maximizes the similarity of the positive pair, which is the embedded patches of $z_i = \text{Encoder} \left( p_i \right)$ and $z_j = \text{Encoder}\left( p_j \right)$. The loss is given by
$l_{i,j}=-\text{sim}(z_i,z_j)/\tau +\log{\sum_{k=1}^{B}\mathbf{1}_{[k\neq i]}\exp(\text{sim}(z_i,z_j)/\tau)}$, where $\text{sim}(\cdot,\cdot)$ is the cosine similarity between two pattern vectors, and $\tau$ is a temperature hyperparameter, following the same approach as in \cite{chen2020simple, chen2020big}.

\subsection{Objectness constraints}\label{subsec:objectness}
Even without the presence of objects, the contrastive training will also memorize common background colors and textures in the form of tight latent distance. This is because (1) they often co-occur with objects as backgrounds, (2) they are sampled frequently during training due to their large spatial span, and quite naturally (3) the positive pair of patches sampled from backgrounds tend to be very similar. Blue skies, green fields or forests, and pavements are typical examples.
 This is not a desirable property for the next step of identifying object boundaries. Our solution is to apply a simple saliency constraint that we call {\it{objectness}} to the contrastive loss so that the training is more sensitive to object-like regions rather than their backgrounds. The constraint is enforced by amplifying the contrastive constraint so that the pairs of patches around objects receive tighter ‘pulls’; the contrastive loss is re-weighed by the average objectness score of the pair.
The objectness is measured based on two properties: (1) typical objects are tight in spatial span and surrounded by background, and (2) common background patterns span much larger areas than objects and these common patterns often occur in many different images. Figure \ref{fig:architecture_modulation} illustrates our approach for measuring the objectness and modulating the contrastive training. We measure the first property for a potential object using the color histogram difference between the object candidate patch and its surrounding background, denoted by $\text{hscore}$. Based on the second property, we developed an effective approach to measure the dissimilarity between a patch and background patterns (which we call '$\text{bscore}$'). Each of these approaches are detailed in the following subsections.

\subsubsection{Histogram distance score}
The bounding box of an object has the general property that the color distributions inside and outside of the box are sufficiently different. Based on this property, we measure the objectness by the color histogram difference between either sides of the bounding box. For a given sampled rectangular patch, a boundary band defined by a larger rectangle with its size proportional to the patch size (green rectangular band in Figure \ref{fig:architecture_modulation}) captures the hypothetical background pixels. Then these background pixels are converted from RGB to HSV to calculate their H-S (2D) histogram. The same method is applied to the pixels inside the inner rectangle (the original patch) to generate another H-S histogram. The difference between these two histograms represents histogram based object saliency, measured by the Hellinger metric. We call this histogram distance '$\text{hscore}$', a histogram based objectness score.

In detail, the $\text{hscore}(p)$ of a patch $p$ is the histogram difference between its inner rectangle $p^I$ and its outer band $p^O$: $\text{hscore}(p) = d_{\text{Hellinger}} \left( h(p^I), h(p^O) \right) - k \cdot \mathbb{E}_{q=q^I\cup q^O, q\in Q}\left[ d_{\text{Hellinger}} \left( h\left( q^I \right), h \left( q^O \right) \right) \right], \label{eqn:dhist}$
where $h(q)$ is the 2D histogram of the patch $q$, $Q$ is a collection of all sampled patches. $d_{\text{Hellinger}}(H_1,H_2)$ is the {\it{Hellinger distance}} between two histograms: 
$     d_{\text{Hellinger}}(H_1,H_2) = \sqrt{1 - \frac{1}{\sqrt{\bar{H_1} \bar{H_2} B^2}} \sum_I \sqrt{H_1(I) \cdot H_2(I)}}$, where $H_1(I)$ and $H_2(I)$ are the frequencies of each bin $I$, and $\bar{H_1}$ and $\bar{H_2}$ are histogram means. The {\it{Hellinger distance}} ranges from value $0$ (identical histograms) to $1$ (orthogonal histograms). The subtraction by the mean $E\left[d_{\text{Hellinger}} \left(\cdot, \cdot \right) \right]$ amplifies the effect of modulation. If we use $k=1$, roughly half of the patch pairs $(p_i, p_j)$ that have lower than average Hellinger distance will have $\text{hscore}{\left( p_i, p_j \right)} < 0$, so that they end up being pushed apart by the contrastive loss. We observed that this causes training to be less stable, thus empirically set the factor at $k=0.5$ to make it more stable.
From each sampled pair of patches $p_i$ and $p_j$ for training, we calculate their average $\text{hscore}$ by $\text{hscore}{\left( p_i,p_j \right)} := \frac{1}{2} \left[ \text{hscore}(p_i) + \text{hscore}(p_j) \right]$
and re-weight the contrastive loss by the average histogram score: $
l_{i,j}^{hist} = \text{hscore}{ \left( p_i,p_j \right)} \cdot l_{i,j}$.
%$\mathbb{E}_{q=q^I\cup q^O, q\in\{Q\}}\left[ d_{\text{Hellinger}}\left( h(q^I), h(q^O) \right) \right]$

\subsubsection{Background dissimilarity}
After the $\text{hscore}(\cdot)$ re-weigh the contrastive loss, certain background patterns are still detected as candidate objects. This is because these background patterns may co-occur with other structures (such as the space between objects or boundaries of background patterns) and may produce $\text{hscore}(\cdot)$ scores that is not low enough to be discarded. 
We develop a method that further suppresses these effects by modeling background patterns dominant in a given image set. We identify them by clustering patches in the pixel space, where the patches are randomly sampled from the same image set.
Figure \ref{fig:architecture_modulation} shows the procedure. First, these patches are flattened to vectors in 32x32x3-dimensional pixel space. We apply the K-means algorithm (k=5, empirically determined to capture diverse patterns) to identify the cluster centers as 'typical background patterns'. Then these cluster centers are stored for measuring the $\text{bscore}(p)$ (background similarity score) of any sampled patch $p$.
Given a patch $p$ and the calculated cluster centers $c_i$ of all sampled patches (flattened to vectors by a $\text{vec}\left( \cdot \right)$ function), its background distance score $\text{bscore}(p)$ is defined as the distance to the closest background cluster center (the minimum over $c_i$'s):
$\text{bscore}(p) = \min_{i} \left\| \text{vec}(p) - c_i \right\|$
Then the score is normalized by its maximum score among all sampled patches $Q$: $\text{bscore}^{\text{norm}}(p) = \frac{\text{bscore}(p)}{\text{maxscore}}$, where $\text{maxscore} = \max_{p \in Q} \text{bscore}(p)$.

Finally, the combination of both the \text{hscore} and the \text{bscore}, by a function $g(\cdot)$, is used to modulate the contrastive loss:
$l_{i,j}^{\text{hist}} = g\left( \text{hscore}{(p_i,p_j)}, \text{bscore}{(p_i,p_j)} \right) \cdot l_{i,j}$.
We empirically determined that a linear combination of these measures $g(a,b) = k_1 \cdot a + k_2 \cdot b$ (with constants $k_1$ and $k_2$) is effective.

\section{Object extraction based on clustering pattern vectors}
The final step for identifying object boundaries is to cluster the learned pattern vectors, where the clustering is performed over the collection of all pattern vectors sampled from every image in the dataset. The Euclidean distance of a pattern vector to the cluster center, denoted by $\text{lscore}$, is the key measure of the presence of an object within the corresponding patch. We also found it beneficial to add both the histogram distance based objectness score $\text{hscore}$ and the background similarity score $\text{bscore}$ to the cluster distance. We call this P-O ({\it{Post-Objectness}}) constraint and we will validate its usefulness in the experimental sections. Here is the summary of the procedure:
\begin{algorithm}
\DontPrintSemicolon
  %\KwInput{}
  %\KwData{Testing set $x$}
  \footnotesize %\scriptsize %\small %\footnotesize	%\scriptsize %\tiny
  1. Fixed number $(200)$ of patches are sampled from each image in the dataset, and combined into a single pool of patches.
  
  2. The trained pattern space embedding maps all the patches to pattern vectors.

  3. 1-mean clustering determines their center and each vector's distance $\text{lscore}$ to the center.
  
  4. P-O (Post-Objectness) constraint: Both the $1-\text{hscore}$ and $1-\text{bscore}$ are calculated and added to the $\text{lscore}$ to calculate the final score: $\text{score} = \text{lscore} + \alpha_h \cdot \left(1 - \text{hscore} \right) + \alpha_b \cdot \left(1 - \text{bscore}\right)$, where $\alpha_h$ and $\alpha_b$ are constants (empirically set as $\alpha_h = \alpha_b = \text{average}\left( \text{lscore} \right) \approx 5.0$).
   
  5. For each image, the patches that output lower scores are selected as object candidates. $N_{\text{candidate}} = 20$ low score patches are selected per image after being sorted by the scores.
   
  6. Non-maxima suppression is applied to the selected candidate patches and final patches are selected as the discovered objects, with the overlap threshold of $0.5$. Maximum of $N_{\text{predict}} = 5$ patches per image are selected as object candidates.
   
    % \KwOutput{}
\caption{Object extraction}
\end{algorithm}

% % % % % % % % % % % % % % % % % % % % 
\section{Experimental results}
\subsection{Datasets and visual results}
\begin{figure}
    \begin{subfigure}{\linewidth}
    \centering
    {(a)}{\includegraphics[width=.9\linewidth,height=0.175\linewidth]{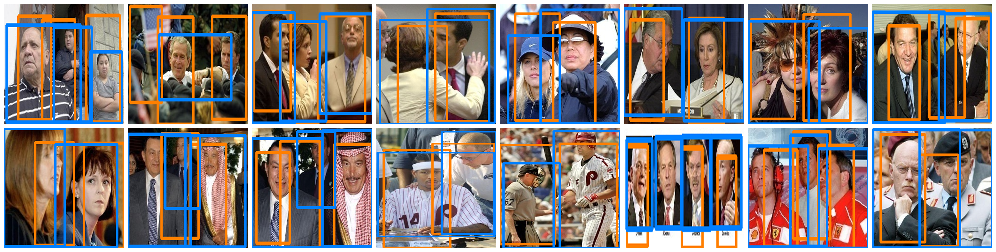}}
    %\caption{FDDB1000}
    \label{fig:fddb100_select_out}
    \end{subfigure}
     \begin{subfigure}{\linewidth}
    \centering
    {(b)}{\includegraphics[width=.9\linewidth,height=0.175\linewidth]{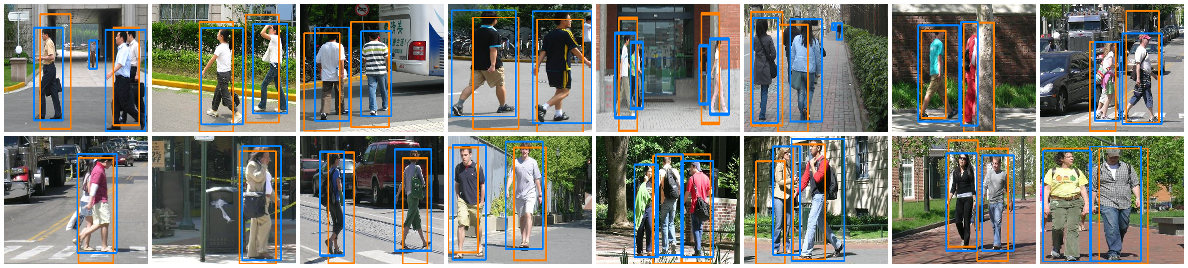}}
    %\caption{PennFudan}
    \label{fig:pennfudan_select_out}
    \end{subfigure}
    \begin{subfigure}{\linewidth}
    \centering
    {(c)}{\includegraphics[width=.9\linewidth,height=0.175\linewidth]{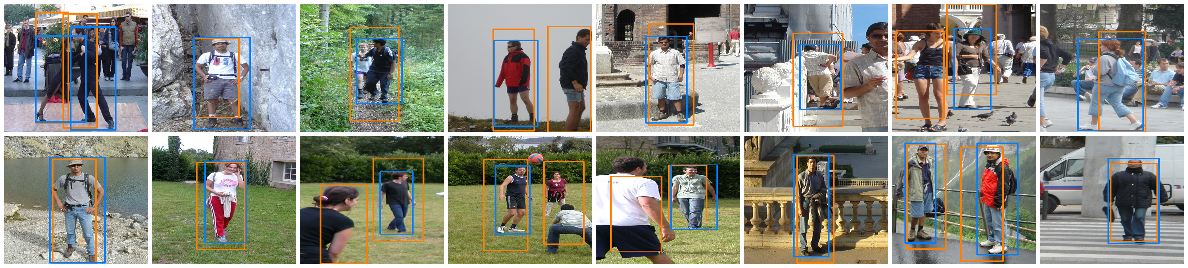}}
    %\caption{INRIA-EZ}
    \label{fig:inria_ez_select_out}
    \end{subfigure}
    \caption{Object discovery results to (a) FDDB100, (b) Penn-Fudan, and (c) INRIA-EZ datasets}
    %  \centering
    %  \subfloat[a]{\includegraphics[width=.99\linewidth]{figures/inria_ez_select_out.r2.png}\label{fig:inria_ez_select_out}}
    %  \centering
    %  \subfloat[b]{\includegraphics[width=.99\linewidth]{figures/pennfudan_select_out.r2.png}\label{fig:pennfudan_select_out}}
    %  \caption{Object discovery results to INRIA-EZ (top) and PennFudan (right) datasets}
     \label{fig:all_select_out}
\end{figure}

\textbf{FDDB-100 dataset}\footnote{\url{http://vis-www.cs.umass.edu/fddb/}} is a subset of the LFW dataset\footnote{\url{http://vis-www.cs.umass.edu/lfw/}} that consists of 5,171 annotated faces in a set of 2,845 images, constructed for face detection research. We selected the first 100 images from FDDB to test the object discovery performance. We do not use the original facial bounding box annotations, but person level annotations as ground truth. This is because our approach picks up any common visible patterns in the image, and that includes body portions outside of faces. Figure \ref{fig:all_select_out} shows the images and the discovered bounding boxes (color coded in orange) compared to the ground truth (color coded in blue). The bounding boxes are not localized with high accuracy, but able to resolve multiple people based on object (person) representation learned from merely 100 images. 

\textbf{Penn-Fudan dataset}\footnote{\url{https://www.cis.upenn.edu/~jshi/ped_html/}} consists of pedestrian images taken around campus and urban streets. It has 170 images with 345 labeled pedestrians, among which 96 images are taken from around University of Pennsylvania, and other 74 are taken from around Fudan University. Figure \ref{fig:all_select_out} shows the images and the ground truth \& discovered bounding boxes marked with the same color scheme.

\textbf{INRIA-EZ dataset} is a subset of the well-known INRIA person dataset\footnote{\url{http://pascal.inrialpes.fr/data/human/}}. The INRIA dataset consists of 902 images (both Train and Test images) that contain images of people against diverse scenes such as streets, tourist attractions, parks, beaches, mountains, trade shows, etc.  We have selected images and objects that are less challenging for detection. The selection criteria are: (1) image having five or less people and (2) ground truth bounding boxes larger than 10K pixels (roughly 60x180 bounding box) within these selected images. These two criteria led to 205 images with 272 ground truth bounding boxes of people. Figure \ref{fig:all_select_out} shows the images and object discovery results. (Object discovery outputs from all images from three datasets are in the supplementary material.)

%\begin{figure}[h]
%\centerline{\includegraphics[width=0.99\textwidth]{figures/fddb100_select_out.3.png}}
%\caption{Face detection results to FDDB-100 images}
%\label{fig:fddb100_select}
%\end{figure}

\subsection{Metric evaluation results}
\begin{table}[h]
\caption{Metric evaluation results for $\text{iou}_\text{thres} = 0.5$}
\setlength{\tabcolsep}{2pt}
{\scriptsize
\begin{tabular}{cccccccccccccc}\toprule
\multicolumn{2}{c}{Modes} & \multicolumn{4}{c}{FDDB100} & \multicolumn{4}{c}{PennFudan}  & \multicolumn{4}{c}{INRIA-EZ}
\\\cmidrule(lr){1-2}\cmidrule(lr){3-6}\cmidrule(lr){7-10}\cmidrule(lr){11-14}
Modulation & Post-Obj  & Corloc & Recall & Precision & F1 & Corloc & Recall & Precision & F1 & Corloc & Recall & Precision & F1\\
x & x   & 70.30 & 53.10 & 54.33 & 53.70  & 63.12 & 35.56 & 23.08 & 27.99  & 55.17 & 51.91 & 17.54 & 26.22\\
 &   & $\pm$2.15 & $\pm$2.31 & $\pm$2.65 & $\pm$2.42  & $\pm$2.11 & $\pm$1.56 & $\pm$1.08 & $\pm$1.26 & $\pm$2.71 & $\pm$2.79 & $\pm$0.97 & $\pm$1.44\\
x & \checkmark & 70.60 & 54.21 & 56.08 & 55.12 & 69.76 & 38.63 & 27.33 & 32.01 & 56.39 & 52.39 & 22.65 & 31.62\\
 &  & $\pm$2.54 & $\pm$2.90 & $\pm$3.00 & $\pm$2.89 & $\pm$3.64 & $\pm$1.58 & $\pm$1.10 & $\pm$1.27 & $\pm$3.46 & $\pm$3.31 & $\pm$1.53 & $\pm$2.10\\
\checkmark & x & 70.20 & 54.00 & 55.05 & 54.51 & 69.00 & 37.21 & 23.99 & 29.17 & 57.85 & 54.26 & 19.09 & 28.24\\
 &  & $\pm$3.19 & $\pm$3.52 & $\pm$3.78 & $\pm$3.60 & $\pm$2.67 & $\pm$0.96 & $\pm$0.81 & $\pm$0.88 & $\pm$4.04 & $\pm$3.89 & $\pm$1.39 & $\pm$2.04\\
\checkmark & \checkmark & 70.30 & 54.21 & 54.70 & 54.44  & 74.00 & 39.69 & 29.23 & 33.66 & 51.71 & 47.48 & 27.64 & 34.94\\
 & & $\pm$2.90 & $\pm$2.21 & $\pm$1.89 & $\pm$1.98 & $\pm$3.67 & $\pm$1.85 & $\pm$1.25 & $\pm$1.48 & $\pm$2.48 & $\pm$2.07 & $\pm$1.23 & $\pm$1.54\\
\bottomrule
\end{tabular}}
\label{table:all_evaluation_out_05}
\end{table}
%\subsubsection{Performance metric for object discovery}
We use two performance metrics for measuring the success of object discovery. For evaluating single-object discovery, we use a metric frequently used in similar object discovery contexts \cite{Cho2015,Vo2019,Vo2020} called correct localization (CorLoc), which is defined as the percentage of images that our model correctly localized at least one target object. The localization accuracy is measured by the common IoU (intersection over union) computation. An IoU over 0.5 ($\text{iou}_\text{thres} = 0.5$) is regarded as success in most of our evaluations, and we consider a relaxed criterion of $\text{iou}_\text{thres} = 0.4$ as well.
For measuring accuracy of multi-object discovery, traditional detection metrics based on Precision and Recall scores also apply. However, we found that AP (Average Precision) is not suitable. In supervised learning scenarios, the detection scores are estimated from the ground truth object presence in each image and at each position. Therefore they provide a decision score consistent across images and within each image. However, our measure of object presence based cluster distance lacks the same consistency and does not interpret directly as a detection score. We use the F1 score for combining recall and precision for our evaluation. The F1 score is developed for measuring the success of information retrieval, and finding an object among randomly generated patches can be regarded as a retrieval problem. Just as the AP is averaged over the range of recall values by varying the detection score threshold, we can control the number of maximum predictions per image to sweep the (recall, precision) ranges. Then we identify the maximum of F1 scores over the range of number of max predictions (from 1 to 5) as the reported accuracy. We also report CorLoc, recall, and precision scores measured at the maximum F1 score. Because the candidate patches are randomly generated, the metric numbers are averaged over ten iterations and their error bounds in terms of standard deviations are listed as $\pm$STD.

The first set of experiments confirm the benefits of the contrastive loss modulation and the P-O constraints. Table \ref{table:all_evaluation_out_05} summarizes the accuracy metrics over different combinations of these features. For the two pedestrian datasets, the modulated contrastive training combined with the P-O constraints (both the histogram and background constraints) achieves the overall best performance compared to other modes. The modulated contrastive training (third and fourth rows vs. first and second rows) clearly demonstrates that the loss modulation improves accuracy regardless of the P-O constraints. On the other hand, the effect of the P-O constraint (first vs. second and third vs. fourth) appears more significant. The FDDB-100 dataset seems to enjoy less benefit from either features. We suspect this is due to many images occupied mostly by faces or bodies (often overlap with other people) without clear boundaries against the background. However, these features do not hurt the accuracy either. We conclude that the combination of both the modulated training and the P-O constraint achieves overall the best accuracy.
\begin{table}[h]
\caption{Metric evaluation results for $\text{iou}_\text{thres} = 0.4$}
\setlength{\tabcolsep}{2pt}
{\scriptsize
\begin{tabular}{cccccccccccccc}\toprule
\multicolumn{2}{c}{Modes} & \multicolumn{4}{c}{FDDB100} & \multicolumn{4}{c}{Penn-Fudan}  & \multicolumn{4}{c}{INRIA-EZ}
\\\cmidrule(lr){1-2}\cmidrule(lr){3-6}\cmidrule(lr){7-10}\cmidrule(lr){11-14}
Modulation & Post-Obj  & Corloc & Recall & Precision & F1 & Corloc & Recall & Precision & F1 & Corloc & Recall & Precision & F1\\
x & \checkmark & 88.20 & 72.69 & 74.81 & 73.73 & 86.47 & 53.92 & 38.23 & 44.74 & 70.39 & 65.65 & 37.49 & 47.72\\
 &   & $\pm$1.47 & $\pm$2.21 & $\pm$1.84 & $\pm$1.91 & $\pm$2.10 & $\pm$2.07 & $\pm$1.75 & $\pm$1.90 & $\pm$2.19 & $\pm$1.91 & $\pm$1.08 & $\pm$1.37\\
 \checkmark & \checkmark & 87.80 & 72.69 & 73.25 & 72.96 & 88.35 & 53.59 & 39.22 & 45.29 & 73.37 & 67.30 & 38.99 & 49.38\\
  &    & $\pm$1.72 & $\pm$2.12 & $\pm$3.02 & $\pm$2.44 & $\pm$1.62 & $\pm$1.99 & $\pm$1.00 & $\pm$1.35 & $\pm$2.74 & $\pm$2.28 & $\pm$1.30 & $\pm$1.65\\
\bottomrule
\end{tabular}}
\label{table:all_evaluation_out_04}
\end{table}

The next experiments test a relaxed detection criterion. Our approach does not utilize ground truth bounding boxes, but rather find emergent patterns from images. Therefore the bounding box estimates are often not accurate. We find many cases where bounding boxes appear to originate from true objects, evidenced by many loose bounding boxes around objects when objects are well separated. For this reason, we report the accuracy results based on the matching criteria of $\text{IoU} > 0.4$ as another legitimate success measure to guide our study. The results are reported in Table \ref{table:all_evaluation_out_04}. We only selected (1) contrastive training with P-O constraints and (2) modulated training with P-O constraints, since we confirmed their benefit from the previous experiments. The accuracy numbers show significant gain, and the benefit of the modulated training is confirmed again for the two pedestrian datasets.

We also confirm the effectiveness of the background based objectness toward images having common backgrounds. Table \ref{table:h_vs_b} summarizes accuracy numbers (only F1 for clear comparison) from different modes: P-O is turned on but modulation by histogram only, or background only, or both. Penn-Fudan data is collected from two campuses therefore the background scenes are limited. The F1 score clearly shows that the $\text{bscore}$ based modulation greatly benefits object discovery from Penn-Fudan images.

\begin{wraptable}{l}{5.8cm}
\caption{Modulation by histogram vs background vs both}\label{table:h_vs_b}
\setlength{\tabcolsep}{2pt}
{\scriptsize
\begin{tabular}{cccccccc}\toprule
\multicolumn{2}{c}{Modulation} & \multicolumn{2}{c}{FDDB100} & \multicolumn{2}{c}{Penn-Fudan} &  \multicolumn{2}{c}{INRIA-EZ}
\\\cmidrule(lr){1-2}\cmidrule(lr){3-4}\cmidrule(lr){5-6}\cmidrule(lr){7-8}
Hist.  & Bgnd. & F1 & std & F1 & std & F1 & std\\
x & x &  55.12 & $\pm$2.89 & 32.01 & $\pm$1.27 & 31.62 & $\pm$2.10\\
\checkmark & x & 58.31 & $\pm$1.91 & 32.77 & $\pm$1.70 & 34.12 & $\pm$1.16\\
x & \checkmark & 56.62 & $\pm$3.90 & 34.44 & $\pm$1.25 & 34.62 & $\pm$1.71\\
\checkmark & \checkmark & 54.44 & $\pm$1.98 & 33.66 & $\pm$1.48 & 34.94 & $\pm$1.54\\
\bottomrule
\end{tabular}
}
\end{wraptable}

\subsection{Latent neighbor analysis}
\begin{figure}
    \begin{subfigure}{\linewidth}
    \centering
    {(a)}{\includegraphics[width=.9\linewidth,height=0.095\linewidth]{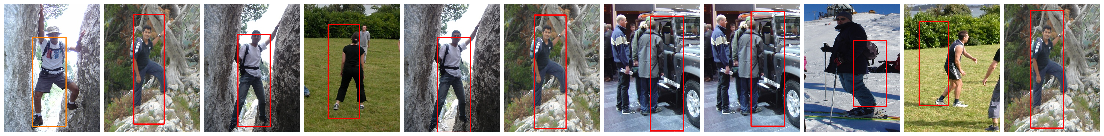}}
    %\caption{}
    \label{fig:nbhd_inria_1}
    \end{subfigure}
    \begin{subfigure}{\linewidth}
    \centering
    {(b)}{\includegraphics[width=.9\linewidth,height=0.095\linewidth]{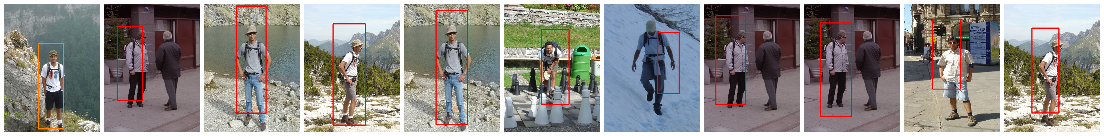}}
    %\caption{}
    \label{fig:nbhd_inria_2}
    \end{subfigure}
    \begin{subfigure}{\linewidth}
    \centering
    {(c)}{\includegraphics[width=.9\linewidth,height=0.095\linewidth]{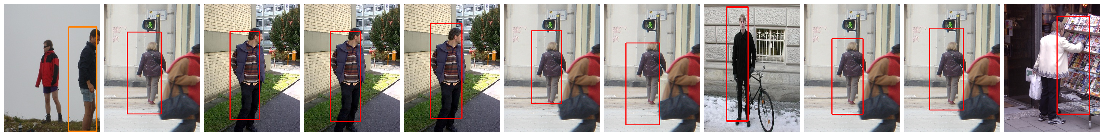}}
    %\caption{}
    \label{fig:nbhd_inria_3}
    \end{subfigure}
    \begin{subfigure}{\linewidth}
    \centering
    {(d)}{\includegraphics[width=.9\linewidth,height=0.095\linewidth]{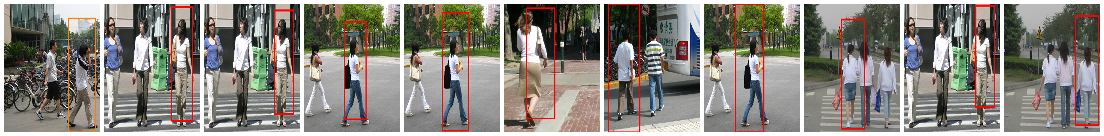}}
    %\caption{}
    \label{fig:nbhd_pennfudan_1}
    \end{subfigure}
    \begin{subfigure}{\linewidth}
    \centering
    {(e)}{\includegraphics[width=.9\linewidth,height=0.095\linewidth]{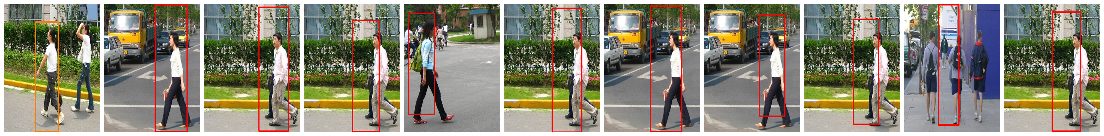}}
    %\caption{}
    \label{fig:nbhd_pennfudan_2}
    \end{subfigure}
     \caption{Discovered objects and their closest neighbors in latent space}
     \label{fig:inria_ez_latent_nbhd}
\end{figure}

We have verified that the contrastive training modulated by objectness constraints leads to a tight cluster. The cluster consists of pattern vectors that encode dominant object-like patterns in a given image set. We further investigated how the pattern space representation encodes more detailed visual features.
Figure \ref{fig:inria_ez_latent_nbhd} illustrates that the trained pattern space captures such visual features based on the nearest neighbors of discovered objects. The first column lists five of the discovered object patches (orange rectangles), and to their right each row shows ten nearest neighbor patches from different images. Rows a and b appear to show common background textures while all rows show consistent body poses. The nearest neighbors also capture similar clothing (c, d, and e) and backpacks people wear (b). However, the analysis is only based on distance, and we have not developed further training constraints or analysis tools to enforce and discover disentangled visual features.

%\subsection{Object Discovery dataset -- Comparison to }
%The Object Discovery dataset \cite{rubinstein2013unsupervised} is widely used as a benchmark dataset in other object discovery work \cite{Cho2015,Vo2019,Vo2019}. The dataset consists of airplane, car, and horse images as well as outlier images without these objects. We use the 100 image subsets for each category as used in the prior studies. 

\section*{Conclusion}
We demonstrated that self-supervised training of randomly generated patches modulated by objectness constraints finds common object patterns from relatively small number of images quite effectively. One of its major limitations is the low localization accuracy and this topic is our active focus of investigation. Another desirable feature is to identify multiple classes of objects from a single unsupervised training. We suspect that combination with Deep Clustering approaches \cite{caron2018deep,Ji_2019_ICCV,vincent2008extracting} may enable multi-class object discovery based on contrastive training. Another avenue is to make the proposed object discovery more practical by incorporating minimal supervision.

\bibliographystyle{apalike}
\bibliography{main_2021_arxiv}

%\medskip
%\small

\end{document}